%% file: main.tex
\documentclass[10pt]{article}

\usepackage[preprint]{tmlr}

\input{math_commands.tex}

\input{mathdef}
\usepackage{hyperref}
\usepackage{url}
\usepackage{graphicx, color}

\usepackage[noorphans]{quoting}
\usepackage[square,sort,comma,numbers]{natbib}

\usepackage{amsmath}
\usepackage{amssymb}
\usepackage{mathtools}
\usepackage{amsthm}
\usepackage{bbm}
\usepackage{booktabs}
\usepackage{subcaption}
\usepackage{wrapfig}

\usepackage{caption}

\usepackage[most]{tcolorbox}

\usepackage[ruled, algo2e, vlined]{algorithm2e}
\SetKwInput{KwInput}{Input}
\SetKwInput{KwOutput}{Output}

\theoremstyle{plain}
\newtheorem{theorem}{Theorem}[section]
\newtheorem{proposition}[theorem]{Proposition}

\newtheorem{corollary}[theorem]{Corollary}
\theoremstyle{definition}

\theoremstyle{remark}

\allowdisplaybreaks

\usepackage{listings}
\usepackage{xcolor}

\definecolor{codegreen}{rgb}{0,0.6,0}
\definecolor{codegray}{rgb}{0.5,0.5,0.5}
\definecolor{codepurple}{rgb}{0.58,0,0.82}
\definecolor{backcolour}{rgb}{1,1,1}

\lstdefinestyle{mystyle}{
    backgroundcolor=\color{backcolour},   
    commentstyle=\color{codegreen},
    keywordstyle=\color{magenta},
    stringstyle=\color{codepurple},
    basicstyle=\ttfamily\footnotesize, % Font size
    breakatwhitespace=false,         
    breaklines=true,
    captionpos=b,
    keepspaces=true,
    numbers=none,
    showspaces=false,
    showstringspaces=false,
    showtabs=false,
    tabsize=4,
    frame=lines      % Add lines at top and bottom
}

\title{Fast EXP3 Algorithms}

\author{\name Ryoma Sato \email rsato@nii.ac.jp \\
  \addr National Institute of Informatics
  \AND
  \name Shinji Ito \email shinji@mist.i.u-tokyo.ac.jp \\
  \addr The University of Tokyo and RIKEN
}

\begin{document}

\maketitle

\begin{abstract}
We point out that EXP3 can be implemented in constant time per round, propose more practical algorithms, and analyze the trade-offs between the regret bounds and time complexities of these algorithms.
\end{abstract}

\section{Introduction}

The computational complexity of adversarial bandits has been largely overlooked. In fact, in much of the literature, the computational complexity of EXP3~\cite{auer2002nonstochastic} is either left unspecified or stated to be linear in the number of arms per round~\cite{hazan2011simple,koolen2024mltlecture9}.

A notable exception is the work by \citet{chewi2022rejection}, who proposed an algorithm that runs in $O(\log^2 K)$ time for the adversarial bandit problem with $K$ arms, at the cost of worsening the constant factor in the regret bound. However, this represents a suboptimal trade-off between computational time and regret.

First, we point out that a practical implementation of EXP3 is possible in $O(\log K)$ time per round exactly, that is, without degrading the constant factor in the regret bound. This improves upon the algorithm of \citet{chewi2022rejection} in both the order of computational time and the regret constant.

Next, we show that by using an advanced general-purpose data structure~\cite{matias2003dynamic}, it is possible to implement EXP3 in $O(1)$ expected time per round exactly, again, without degrading the constant factor in the regret bound. This also improves upon the algorithm of \citet{chewi2022rejection} in both the order of computational time and the regret constant.

However, this data structure is exceedingly complex.

Therefore, we propose a simpler and more easily analyzable implementation of EXP3 that runs in $O(1)$ expected time per round.

Finally, we analyze the trade-offs between computational time and regret when making EXP3 anytime. We show that the commonly used doubling trick is not optimal and that better trade-offs exist.

We summarize our main results in Table~\ref{tab: main_results}.

\section{Background: Adversarial Bandits and EXP3}

In this section, we define the setting of the adversarial bandit problem addressed in this paper and describe the details of the baseline algorithm EXP3 along with its regret upper bound.

\subsection{Problem Setting}

We consider a slot machine with $K \ge 2$ arms over a game of $T$ rounds.
The adversary (also called the environment) and the player interact according to the following protocol.

In each round $t = 1, \dots, T$:
\begin{enumerate}
    \item \textbf{Adversary's turn:}
    The adversary determines the loss vector $\ell_t = (\ell_{t,1}, \dots, \ell_{t,K}) \in [0, 1]^K$ for each arm based on the history $\mathcal{H}_{t-1} = (a_1, \ell_{1, a_1}, \dots, a_{t-1}, \ell_{t-1, a_{t-1}})$.
    Crucially, while the adversary knows the player's algorithm (policy), it must fix $\ell_t$ before knowing the realization of the player's internal randomness at round $t$ (i.e., before knowing which arm $a_t$ will be selected).

    \item \textbf{Player's turn:}
    The player follows the algorithm and selects one arm $a_t \in \{1, \dots, K\}$ using the observable history and its own internal randomness.

    \item \textbf{Observation and loss:}
    The player observes only the loss $\ell_{t, a_t}$ of the selected arm and incurs this loss. The losses $\ell_{t, i}$ $(i \ne a_t)$ of the unselected arms are not observed (bandit feedback).
\end{enumerate}

In this paper, we primarily consider the setting where the horizon $T$ is known in advance. We will discuss the setting where the horizon is unknown in Section \ref{sec: anytime}.

In this setting, the player's objective is to minimize the expected pseudo-regret $\bar{R}_T$ defined as follows.
\begin{equation}
    \bar{R}_T := \max_{i \in \{1, \dots, K\}} \mathbb{E}\left[ \sum_{t=1}^T \ell_{t, a_t} - \sum_{t=1}^T \ell_{t,i} \right]
\end{equation}
Here, the expectation $\mathbb{E}[\cdot]$ is taken over the randomness in the player's algorithm (and the adversary's randomness if the adversary is stochastic).

\begin{table}[t]
    \centering
    \caption{Summary of main results in this paper. All proposed methods outperform \citet{chewi2022rejection} in both time complexity and regret constant. The regret coefficient represents the coefficient of $\sqrt{K T \ln K}$ in the expected pseudo-regret for the anytime setting; smaller is better.}
    \label{tab: main_results}
    \resizebox{\textwidth}{!}{%
    \begin{tabular}{lcc}
        \toprule
        Algorithm & Time Complexity per Round & Regret Coefficient \\
        \midrule
        Naive implementation & $O(K)$ worst-case & $2$ \\
        \citet{chewi2022rejection} & $O(\log^2 K)$ worst-case processing + $O(\log K)$ expected sampling & $4$ \\
        Binary tree (Section~\ref{sec: log_time_exp3}) & $O(\log K)$ worst-case & $2$ \\
        Advanced data structure (Section~\ref{sec: constant_time_exp3}) & $O(1)$ expected & $2$ \\
        Alias method (Section~\ref{sec: simpler_constant_time_exp3}) & $O(1)$ expected & $2$ \\
        \bottomrule
    \end{tabular}
    }
\end{table}

\subsection{The EXP3 Algorithm}

EXP3~\cite{auer2002nonstochastic} is an algorithm that maintains a weight $w_{t,i}$ for each arm and selects arms according to a probability distribution based on these weights.

We fix the learning rate $\eta := \sqrt{\frac{2\ln K}{K T}}$.
The initial weights are set to $w_{1,i} = 1$ for $i = 1, \dots, K$, and in each round $t$, we define the total weight as $W_t := \sum_{i=1}^K w_{t,i}$.
The player selects arm $a_t$ according to the following probability distribution $p_t = (p_{t,1}, \dots, p_{t,K})$.
\begin{equation}
    p_{t,i} := \frac{w_{t,i}}{W_t}
\end{equation}
After selecting arm $a_t$ and observing loss $\ell_{t, a_t}$, the weights are updated as follows.
\begin{equation}
    w_{t+1,i} :=
    \begin{cases}
         w_{t,i} \exp\left( - \eta \frac{\ell_{t,i}}{p_{t,i}} \right) & (\text{if } i = a_t) \\
         w_{t,i} & (\text{otherwise})
    \end{cases}
    \label{eq: exp3_update_rule}
\end{equation}
This is equivalent to performing an exponential weight update using an inverse probability weighting (IPW) estimator that treats the losses of unobserved arms as zero. When losses are in the range $[0, 1]$, the standard analysis yields the following upper bound on the regret~\cite[Theorem 3.1]{bubeck2012regret}.
\begin{equation}
    \bar{R}_T \le \sqrt{2} \cdot \sqrt{T K \ln K}.
\end{equation}

\subsection{Computational Model}

In this paper, we adopt the RAM model, which assumes that basic arithmetic operations (addition, subtraction, multiplication, division, exponential functions, logarithmic functions, random number generation, etc.) can be executed in $O(1)$ time. This is the model commonly used when dealing with data structures for sampling, such as in~\citet{matias2003dynamic}.

\section{Practical Logarithmic-Time Implementation of EXP3} \label{sec: log_time_exp3}

\begin{figure}[t]
    \centering
    \includegraphics[width=0.8\textwidth]{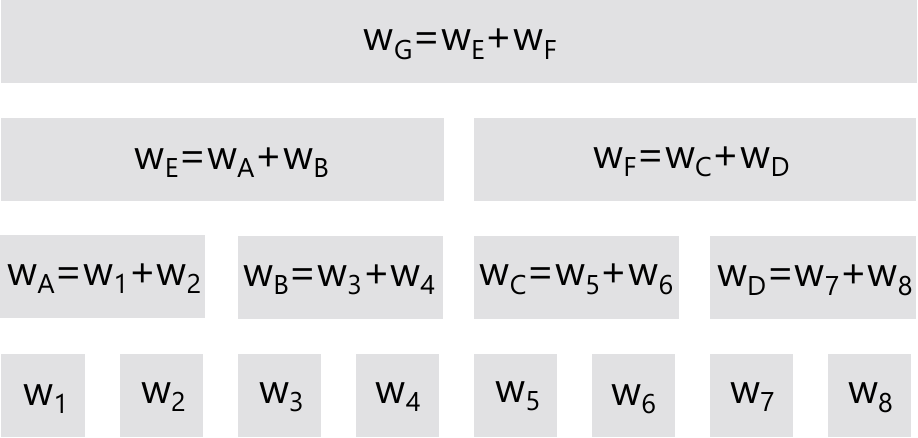}
    \caption{Example of a segment tree managing EXP3 weights. Each leaf stores the weight of an arm, and each internal node stores the sum of its children's weights.}
    \label{fig:segment_tree}
\end{figure}

We demonstrate that a practical implementation of EXP3 is possible in $O(\log K)$ time per round exactly, that is, without degrading the constant factor in the regret bound.

The key data structure is a static balanced tree (e.g., segment tree, B-tree). This tree has $K$ leaves, where each leaf $i \in \{1, \dots, K\}$ stores the current weight $w_{t,i}$. Additionally, each internal node $v$ stores the sum of the weights of its child nodes (Figure~\ref{fig:segment_tree}).

\begin{figure}[t]
    \centering
    \includegraphics[width=0.8\textwidth]{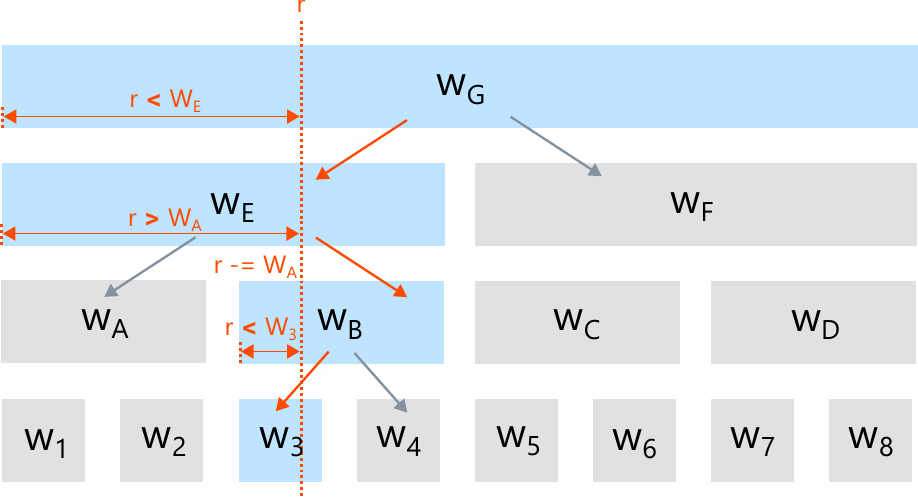}
    \caption{Visualizing the sampling process as a hierarchical interval search. A single random value $r \in [0, W_t)$ (dashed red line) is generated. The algorithm identifies the leaf node covering $r$ by traversing down the tree in $O(\log K)$ time.}
    \label{fig:segment_tree_sampling}
\end{figure}

Arm sampling in EXP3 at round $t$ can be performed as follows (Figure~\ref{fig:segment_tree_sampling}).

\begin{enumerate}
    \item Generate a random number $r$ uniformly in $[0, W_t)$, where $W_t$ is the total weight stored at the root.
    \item Place a pointer $v$ at the root of the tree.
    \item While $v$ is an internal node:
    \begin{enumerate}
        \item Let $W_l$ denote the value of the left child of the current node $v$.
        \item If $r < W_l$, move the pointer to the left child.
        \item Otherwise, subtract $W_l$ from $r$ and move the pointer to the right child.
    \end{enumerate}
    \item The selected arm corresponds to the leaf node $v$.
\end{enumerate}

\begin{figure}[t]
    \centering
    \includegraphics[width=0.8\textwidth]{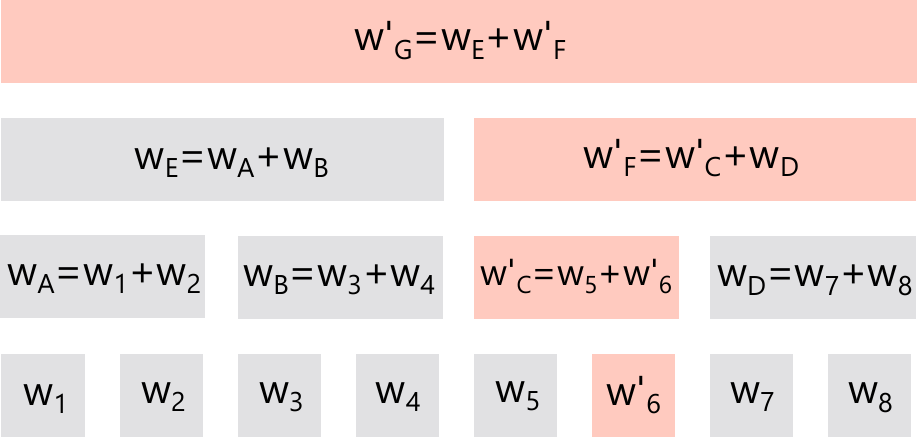}
    \caption{Example of weight update using a segment tree. By propagating updates from the leaf to the root, the weights can be updated in $O(\log K)$ time.}
    \label{fig:segment_tree_update}
\end{figure}

The weight update phase in EXP3 can also be performed as follows (Figure~\ref{fig:segment_tree_update}).

\begin{enumerate}
    \item Update the value of the leaf corresponding to the updated arm.
    \item Move up to the parent node and set its new value as the sum of its children's new values.
    \item Repeat this process until reaching the root.
\end{enumerate}

Each round performs (i) one weighted sampling query by descending the tree and (ii) one point update at the sampled leaf followed by recomputation along the root path; both operations take $O(\log K)$ worst-case time. Consequently, EXP3 can be implemented \emph{exactly} with $O(\log K)$ worst-case time per round without changing its regret guarantee.

\begin{figure}[t]
\centering
\begin{lstlisting}[language=Python, caption={Python implementation of Log-Time EXP3 using Segment Tree. Indices $1$ through $M-1$ represent internal nodes, and $M$ through $2M-1$ represent leaves. The left child of index $i$ is $2i$, and the right child is $2i+1$. Conversely, the parent of index $i$ is $\lfloor i/2 \rfloor$. For illustration purposes, we use raw values; in practice, computations in the log domain are necessary to avoid underflow and ensure numerical stability.}, label={lst:exp3_segtree}]
# Initialization
M = 1
while M < K: M *= 2      # Padding to power of 2
tree = np.zeros(2 * M)   # Segment tree array
tree[M:M+K] = 1.0        # Initialize leaves (weights)
for i in range(M - 1, 0, -1):
    tree[i] = tree[2*i] + tree[2*i+1] # Build tree

for t in range(T):
    # 1. Sampling (Time O(log K))
    idx = 1
    val = np.random.rand() * tree[1] # Random value in [0, W_t)
    while idx < M:
        if val < tree[2*idx]:        # Go left
            idx = 2 * idx
        else:                        # Go right
            val -= tree[2*idx]
            idx = 2 * idx + 1
    arm = idx - M

    # 2. Observation and Estimation
    loss = observe(arm)
    prob = tree[idx] / tree[1]   # p_{t,i} = w_{t,i} / W_t
    est_loss = loss / prob

    # 3. Weight Update (Time O(log K))
    tree[idx] *= np.exp(-eta * est_loss)
    while idx > 1:                   # Propagate up
        idx //= 2
        tree[idx] = tree[2*idx] + tree[2*idx+1]
\end{lstlisting}
\end{figure}

Code~\ref{lst:exp3_segtree} shows the Python implementation. It is remarkably simple. The only function that needs to be provided is \texttt{observe}. The \texttt{observe} function takes an arm index (0-indexed) and returns the loss of that arm.

A notable practical advantage of this implementation is that it requires generating only a single random number per round. This is in contrast to the algorithm of \citet{chewi2022rejection} and the methods discussed in Sections \ref{sec: constant_time_exp3} and \ref{sec: simpler_constant_time_exp3}, which typically require multiple random samples. Given that random number generation is often computationally expensive, minimizing these calls offers a significant speed advantage. Furthermore, static balanced trees such as segment trees and B-trees are known for their excellent cache efficiency, making them not only theoretically efficient but also fast in practice. In reality, arithmetic operations and memory accesses may have different execution times. When memory access is costly, increasing the branching factor $B$ of a B-tree to reduce its height can lead to faster execution.

\section{Constant-Time EXP3} \label{sec: constant_time_exp3}

By using more advanced data structures, constant-time execution is theoretically possible.

The data structure of \citet{matias2003dynamic} maintains a sequence $w_1, \ldots, w_K$ and supports the following operations in $O(1)$ expected time.
\begin{itemize}
    \item Modify the value of weight $w_i$.
    \item Sample index $i$ with probability $\frac{w_i}{\sum_j w_j}$.
\end{itemize}

Using this data structure, exact EXP3 can be executed in constant time. Computing the IPW estimator requires calculating $\frac{w_{t,i}}{W_t}$, which can be done by separately maintaining the total weight $W_t$.

However, this data structure is extremely complex to implement, as it requires precomputing all results for subproblems of size $O(\log \log K)$ and preparing lookup tables of size $(\log K)^{O(\log \log K)}$. To the best of our knowledge, it is rarely used in practice.

\section{A Simpler Constant-Time EXP3 Using the Alias Method} \label{sec: simpler_constant_time_exp3}

In this section, we propose a simpler constant-time algorithm specialized for EXP3. The main building block is the alias method, which we describe below.

\subsection{Background: The Alias Method}

\begin{figure}[t]
    \centering
    \includegraphics[width=0.6\textwidth]{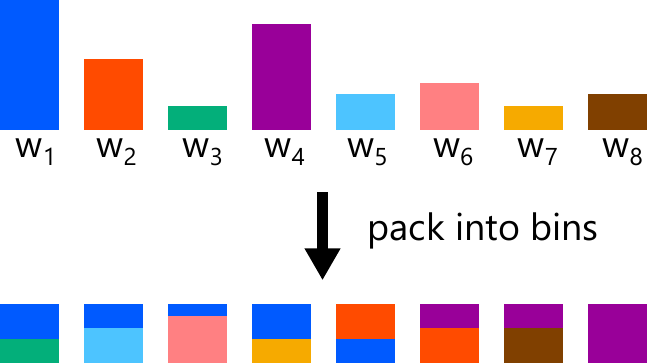}
    \caption{Example of the alias method data structure. Each bin contains at most two elements. By uniformly selecting a bin at random and then selecting an element within the bin with probability proportional to its weight, sampling from the original weight distribution becomes possible.}
    \label{fig:alias_structure}
\end{figure}

The alias method~\cite{walker1977efficient} is a data structure that takes a sequence $w_1, \dots, w_K$ and, after $O(K)$ preprocessing time, enables sampling index $i$ with probability $\frac{w_i}{\sum_k w_k}$ in $O(1)$ worst-case time. Note that the alias method is a static data structure and does not support dynamic updates.

The preprocessing of the alias method proceeds as follows.

First, compute the average weight $\bar{W} = \frac{\sum_k w_k}{K}$. For each $w_i$, assign it to the Small group if $w_i < \bar{W}$, or to the Large group if $w_i \ge \bar{W}$.

Next, repeat the following operation $K$ times (for $i = 1, \dots, K$):
\begin{enumerate}
    \item Take an element $s$ from the Small group and an element $l$ from the Large group.
    \item Pack element $s$ with weight $w_s$ and element $l$ with weight $\bar{W} - w_s$ into the $i$-th ``bin.''
    \item Update the remaining weight of the Large element as $w_l \leftarrow w_l - (\bar{W} - w_s)$.
    \item If the updated $w_l$ is less than $\bar{W}$, return it to the Small group; otherwise, return it to the Large group.
\end{enumerate}
After preprocessing, we obtain $K$ bins, each with capacity $\bar{W}$.
Each bin contains at most two elements (Figure~\ref{fig:alias_structure}).

Using this structure, sampling an element proportional to the original weights can be performed in the following two steps.

\begin{enumerate}
    \item \textbf{Bin selection:} First, sample a bin uniformly at random, selecting bin $i$.
    \item \textbf{Element selection:} Let the contents of the selected bin $i$ be $(s, w_s), (l, w_l')$ where $w_s + w_l' = \bar{W}$.
    Select $s$ with probability $\frac{w_s}{\bar{W}}$ and $l$ with probability $\frac{w_l'}{\bar{W}}$.
\end{enumerate}

This enables sampling from a probability distribution proportional to the weights (areas). Since each step can be executed in $O(1)$ time, the overall sampling time is $O(1)$. Although there are $K$ bins, uniform sampling over bins and the fact that each bin contains only two elements make this efficient.

\subsection{Proposed Method: Constant-Time EXP3 Using the Alias Method}

By introducing the following two ideas to the alias method, we achieve EXP3 in expected $O(1)$ time.

\begin{enumerate}
    \item \textbf{Periodic reconstruction:} Rebuild the alias method data structure from scratch every $K$ rounds.
    \item \textbf{Rejection sampling:} In each round, sample an arm from the alias method data structure built at the previous checkpoint, and accept that arm with a probability based on the current weights. If rejected, repeat the sampling.
\end{enumerate}

Code~\ref{lst:exp3_python} shows the Python code. It is remarkably simple. The only components that need to be provided are the \texttt{Alias} class and the \texttt{observe} function. The \texttt{Alias} class represents the alias method data structure and can be replaced with, for example, \texttt{vose.Sampler} from the \texttt{vose} package. The \texttt{observe} function takes an arm index (0-indexed) and returns the loss of that arm.

\begin{figure}[t]
\centering
\begin{lstlisting}[language=Python, caption={Python implementation of Constant-Time EXP3. For illustration purposes, we use raw values; in practice, computations in the log domain and/or periodic renormalization are necessary to avoid underflow and ensure numerical stability.}, label={lst:exp3_python}]
# Initialization
w = np.ones(K)            # Current weights w_t
W = float(K)              # Total weight W_t
w_snap = w.copy()         # Snapshot weights w_tau
sampler = Alias(w_snap)   # Alias Table (Construction O(K))

for t in range(T):
    # 1. Sampling (Expected time O(1))
    while True:
        k = sampler.sample()             # Sample candidate k in O(1)
        accept_prob = w[k] / w_snap[k]
        if np.random.rand() < accept_prob:
            arm = k
            break

    # 2. Observation and Estimation
    loss = observe(arm)                  # Observe loss
    prob = w[arm] / W                    # Selection probability p_t
    est_loss = loss / prob               # IPW estimator

    # 3. Weight Update (Strictly O(1))
    w_old = w[arm]
    w[arm] = w_old * np.exp(-eta * est_loss)  # Decrease weight
    W -= (w_old - w[arm])                     # Update total weight

    # 4. Periodic Reset (Amortized O(1))
    if (t + 1) % K == 0:
        w_snap = w.copy()         # Update snapshot
        sampler = Alias(w_snap)   # Reconstruct Alias Table
\end{lstlisting}
\end{figure}

Through detailed analysis, we can show that sampling can be performed in expected $O(1)$ time. Additionally, the reconstruction of the alias method takes $O(K)$ time every $K$ rounds. Therefore, we can execute EXP3 in $O(K)$ time per $K$ rounds, which is $O(1)$ expected amortized time per round. Furthermore, as we will discuss in Section \ref{sec: to_expected_time}, the expected amortized time complexity can be strengthened to expected time complexity in a stronger sense.

\subsection{Analysis}

\textbf{Assumption.} We set the learning rate to $\eta := \sqrt{\frac{2\ln K}{K T}}$ as in standard fixed-horizon EXP3. The acceptance rate of the rejection sampler depends on how much the total weight $W_t$ can decrease between two reconstructions. To obtain a clean universal constant, we focus on the nontrivial regime $T \ge 2K\ln K$, which implies $\eta K \le 1$ (and in particular $\eta \le 1/2$ for $K\ge 2$). Note that when $T < 2 K \ln K$, the right-hand side of the EXP3 regret bound $\bar{R}_T \le \sqrt{2} \cdot \sqrt{T K \ln K}$ exceeds $T$, making it almost meaningless to run EXP3. Therefore, the assumption $T \ge 2 K \ln K$ is reasonable. Outside this regime, our algorithm is still well-defined, but the constant in the expected sampling time may deteriorate.

\textbf{Notation.} Let $w_{t, 1}, \dots, w_{t, K}$ denote the weight of each arm at time $t$, and let $W_t = \sum_{j=1}^K w_{t, j}$ denote the total weight. Let $\tau$ be the round just before which the alias method data structure was last reconstructed.

\begin{proposition} \label{prop: simplified-alias-properties}
    The expected number of sampling attempts required until an arm is accepted in rejection sampling is at most $e^2 \le 7.39$.
\end{proposition}

\begin{tcolorbox}[colframe=gray!10,colback=gray!10,sharp corners,breakable]
\begin{proof}
    The acceptance probability in rejection sampling is \begin{align}
        \sum_{i = 1}^K p_{\tau,i} \cdot \frac{w_{t,i}}{w_{\tau,i}} &= \sum_{i=1}^K \frac{w_{\tau,i}}{W_{\tau}} \cdot \frac{w_{t,i}}{w_{\tau,i}} \\
        &= \frac{1}{W_{\tau}} \sum_{i=1}^K w_{t,i} \\
        &= \frac{W_t}{W_{\tau}}
    \end{align} Therefore, the expected number of sampling attempts is $\frac{W_{\tau}}{W_t}$. Here, \begin{align}
        \label{eq: bound_Wt1}
        W_{t+1} &= W_t - w_{t,a_t} + w_{t+1,a_t} \\
        &= W_t - w_{t,a_t} + w_{t,a_t} \exp\left( - \eta \frac{\ell_{t,a_t}}{p_{t,a_t}} \right) \\
        &\ge W_t - w_{t,a_t} + w_{t,a_t} \left( 1 - \eta \frac{\ell_{t,a_t}}{p_{t,a_t}} \right) \\
        &= W_t - w_{t,a_t} \eta \frac{\ell_{t,a_t}}{p_{t,a_t}} \\
        &= W_t - \eta w_{t, a_t} \frac{\ell_{t,a_t}}{\frac{w_{t,a_t}}{W_t}} \\
        &= W_t - \eta W_t \ell_{t,a_t} \\
        &\ge W_t - \eta W_t \\
        &= W_t (1 - \eta).
    \end{align} Therefore, \begin{align}
        W_t &\ge W_{\tau} (1 - \eta)^{t - \tau} \\
        &\ge W_{\tau} (1 - \eta)^K \\
        &\stackrel{\text{(a)}}{\ge} W_{\tau} e^{-2 \eta K} \\
        &\stackrel{\text{(b)}}{\ge} W_{\tau} e^{-2}
    \end{align} where (a) follows from $\eta \le \frac{1}{2}$ and $1 - x \ge e^{-2x}$ for $x \in \left[0, \frac{1}{2}\right]$, and (b) follows from $\eta K \le 1$. Therefore, the expected number of sampling attempts is $\frac{W_{\tau}}{W_t} \le e^2$.
\end{proof}
\end{tcolorbox}

\begin{corollary}
    The proposed method can execute EXP3 in expected amortized $O(1)$ time.
\end{corollary}

\begin{tcolorbox}[colframe=gray!10,colback=gray!10,sharp corners,breakable]
\begin{proof}
    By Proposition~\ref{prop: simplified-alias-properties}, arm sampling is possible in expected $O(1)$ time. Computing the IPW estimator can be done in $O(1)$ time by separately maintaining each weight $w_{t,i}$ and the total weight $W_t$ in an array. Since only one arm's weight is updated per round, the weight update also takes $O(1)$ time. Furthermore, the alias method is reconstructed in $O(K)$ time once every $K$ rounds. Therefore, EXP3 can be executed in expected amortized $O(1)$ time per round.
\end{proof}
\end{tcolorbox}

\subsection{Improving Expected Amortized Time to Expected Time} \label{sec: to_expected_time}

\begin{figure}[t]
    \centering
    \includegraphics[width=\textwidth]{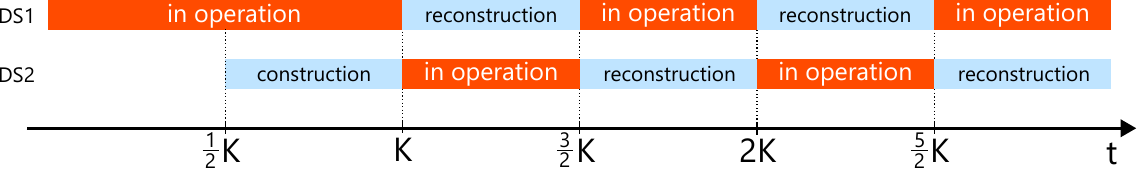}
    \caption{Illustration of double buffering.}
    \label{fig:double_buffering}
\end{figure}

In the above method, a wait of $\Theta(K)$ time inevitably occurs once every $K$ rounds, which is undesirable in online environments where data arrives continuously and predictions must be made immediately.

This can be resolved using double buffering. Specifically, we maintain two data structures: while one is being reconstructed in the background, the other is used for sampling. Once reconstruction is complete, we switch to using the newly reconstructed data structure and begin reconstructing the other one. By interleaving approximately $2$ clock cycles ($O(1)$ time) of background processing per round, reconstruction can be completed in $\frac{K}{2}$ rounds, and the subsequent $\frac{K}{2}$ rounds can perform sampling in $O(1)$ time using the new data structure (Figure~\ref{fig:double_buffering}). This improves the expected amortized time complexity to expected time complexity.

\section{Extension to EXP4}
The approaches given in Sections~\ref{sec: log_time_exp3}, \ref{sec: constant_time_exp3} and \ref{sec: simpler_constant_time_exp3} can also be applied to the {multi-armed bandit with expert advice} problem,
yielding a computationally efficient implementation of the {EXP4} algorithm~\citep{auer2002nonstochastic}.
In this problem setting,
we assume that in each round $t$,
each expert $j \in \{1, \ldots, N\}$ recommend an arm $e_t(j) \in \{1,\ldots, K \}$,
and the learner decides which arm to choose on the basis of these advices.
We consider a computational model in which one has access to an oracle that,
given an expert index $j$,
returns both the arm $e_t(j)$ recommended by expert $j$ and the list $E_t(j) = \{ j' \in \{ 1, \ldots, N \} \mid e_t(j') = e_t(j) \}$ of experts who recommend the same arm.

EXP4 maintains weights $(\tilde{w}_{t,j})_{j=1}^N$ over experts $j \in \{ 1, \ldots, N\}$,
rather than over arms,
and samples an expert $j_t$ following a probability distribution $(\tilde{p}_{t,j})_{j=1}^N$ proportional to the weight $(\tilde{w}_{t,j})_{j=1}^N$.
The weight update rule follows the form of 
$\tilde{w}_{t+1, j} := \tilde{w}_{t, j} \exp\left(-\eta \frac{\mathbb{I}[j \in E_t(j_t)]\ell_{t, e_t(j_t)}}{\sum_{j' \in E_t(j)} \tilde{p}_{t,j'}}\right)$
instead of equation~(\ref{eq: exp3_update_rule}),
and only the weights of experts in $E_t(j_t)$ are updated.
Consequently, the computational cost of each update is bounded by $|E_t(j_t)|$ times the cost in the EXP3 setting.
We here note that,
the bound on the counterpart of $W_{t+1}$ in equation~(\ref{eq: bound_Wt1}) in the proof of Proposition~\ref{prop: simplified-alias-properties} still holds as
\begin{equation}
    \tilde{W}_{t+1}
    :=
    \sum_{j =1}^N \tilde{w}_{t+1, j}
    = 
    \tilde{W}_{t} + \sum_{j \in E_t(j_t)} (\tilde{w}_{t+1,j} - \tilde{w}_{t,j})
    \ge
    \tilde{W}_{t} - \sum_{j \in E_t(j_t)} \tilde{w}_{t,j} \eta \frac{\ell_{t,e_t(j_t)}}{\sum_{j' \in E_t(j)} \tilde{p}_{t, j'}}
    =
    \tilde{W}_{t}(1 - \eta \ell_{t, e_t(j_t)})
\end{equation}
and thus a bound analogous to Proposition~\ref{prop: simplified-alias-properties} also holds.
For further details of the EXP4 algorithm and its regret analysis, we refer the reader to the paper by \citet{auer2002nonstochastic}.

From the above observation,
approaches given in Sections~\ref{sec: constant_time_exp3} and \ref{sec: simpler_constant_time_exp3}
yield implementations of EXP4 with 
the per-round computational complexity of $O( \max_{t, j} \{ |E_t(j)| \}) $ as well as the total computational complexity of $O( \sum_{t=1}^T \max \{ |E_t(j)| \}_{j=1}^N )$.
In the case of the approach in Section~\ref{sec: log_time_exp3}, the complexity increases by an additional $O(\log N)$ factor.

\section{Making EXP3 Anytime} \label{sec: anytime}

Up to this point, we have assumed that the horizon $T$ is known. We now consider the case where the horizon $T$ is unknown. There are primarily three approaches to handle this: the first is discussed in Section \ref{sec: doubling_trick}, while the others involve combining Section \ref{sec: anytime_exp3_ftrl} with either Section \ref{sec: anytime_exp3_rejection_sampling} or Section \ref{sec: anytime_exp3_delayed_updates}.

\subsection{Doubling Trick} \label{sec: doubling_trick}

The simplest approach is the doubling trick. This method divides rounds into blocks of sizes $1, 2, 4, 8, \dots$ that double in length, and runs EXP3 independently on each block. With this method, the upper bound on regret of order $O(\sqrt{T})$ is multiplied by a factor of $(2 + \sqrt{2})$ (see, e.g., Exercise 2.8 of \citep{cesa2006prediction}). For example, since the regret upper bound of fixed-horizon EXP3 is $\sqrt{2} \cdot \sqrt{K T \ln K}$, applying the doubling trick for anytime conversion yields a regret upper bound of $\sqrt{2} (2 + \sqrt{2}) \cdot \sqrt{K T \ln K} \le 4.83 \cdot \sqrt{K T \ln K}$.

\subsection{Time-Varying Parameters via Follow the Regularized Leader} \label{sec: anytime_exp3_ftrl}

EXP3 can be naturally cast into the Follow the Regularized Leader (FTRL) framework, and anytime conversion is possible by using time-varying parameters in FTRL. Specifically, the learning rate is set to \begin{equation}
        \eta_t := \sqrt{\frac{\ln K}{K t}}. \label{eq: anytime-exp3-params}
\end{equation} The initial weights are set to $w_{1,i} = 1$ for $i = 1, \dots, K$, and in each round $t$, we define the total weight as $W_t := \sum_{i=1}^K w_{t,i}$.
The player selects arm $a_t$ according to the following probability distribution $p_t = (p_{t,1}, \dots, p_{t,K})$.
\begin{equation}
    p_{t,i} := \frac{w_{t,i}}{W_t}.
\end{equation}
After selecting arm $a_t$ and observing loss $\ell_{t, a_t}$, the weights are updated as follows.
\begin{equation}
    w_{t+1,i} :=\exp\left( - \eta_t \sum_{s=1}^t \frac{\ell_{s,i} \mathbb{I}[a_s = i]}{p_{s,i}} \right). \label{eq: anytime-exp3-weight-update}
\end{equation}
With this method, the regret upper bound becomes $2 \sqrt{K T \ln K}$.

However, this method cannot be easily accelerated. This is because equation~(\ref{eq: anytime-exp3-weight-update}) requires updating all weights using the current learning rate $\eta_t$. This is in contrast to fixed-horizon EXP3, where only one arm's weight changes. Note that simply making the weight parameter $\eta_t$ in the fixed-horizon EXP3 algorithm time-varying causes past samples to have large weights while recent samples have small weights, which prevents proper control of the anytime regret.

There are two approaches to address this issue.

\subsection{Rejection Sampling} \label{sec: anytime_exp3_rejection_sampling}

The first approach uses rejection sampling. This enables sampling according to $\frac{w_{t,i}}{W_t}$ without explicitly maintaining the exact distribution $\frac{w_{t,i}}{W_t}$. \citet{chewi2022rejection} use this approach.

As a method specialized for anytime EXP3, we can use different learning rates for the target distribution $p$ and the proposal distribution $q$. The target distribution $p$ uses the standard anytime EXP3 parameters as in equation~(\ref{eq: anytime-exp3-params}). On the other hand, the proposal distribution $q$ updates the learning rate every $K$ rounds. Dividing rounds into blocks of $K$ rounds each and letting the last round of the block be $\tau(t) = \lceil \frac{t}{K} \rceil K$, we use $\eta_{\tau(t)}$. Since this is a fixed learning rate within each block, we can apply the logarithmic-time EXP3 of Section~\ref{sec: log_time_exp3} or the constant-time EXP3 of Sections~\ref{sec: constant_time_exp3} and~\ref{sec: simpler_constant_time_exp3} to the proposal distribution $q$. Although the data structure needs to be reconstructed with a new learning rate every $K$ rounds, the per-round computational complexity remains $O(1)$. In particular, Section~\ref{sec: simpler_constant_time_exp3} already reconstructs the data structure every $K$ rounds anyway, so no additional computational cost is incurred. Furthermore, since the proposal distribution $q$ and target distribution $p$ are sufficiently close, the acceptance rate of rejection sampling also remains constant. This enables an anytime constant-time EXP3.

The problem is that we need to explicitly compute $p_{t,i}$ for calculating the IPW estimator. While the proposal distribution value can be computed in constant time, computing the target distribution value---especially its normalizing constant $W_t$---takes $O(K)$ time. To avoid this, we need to estimate $p_{t,i}$, but the variance of this estimator may worsen the regret.

Indeed, \citet{chewi2022rejection} use this approach, but the constant in the regret upper bound doubles compared to the standard anytime EXP3, becoming $4 \sqrt{K T \ln K}$.

\subsection{Delayed Parameter Updates} \label{sec: anytime_exp3_delayed_updates}

The second approach uses delayed parameter updates. Unlike rejection sampling, this approach defines the target distribution itself using delayed parameters. Specifically, we divide update rounds into blocks of $K$ rounds each, and letting the last round of the block be $\tau(t) = \lceil \frac{t}{K} \rceil K$, we use $\eta_{\tau(t)}$ as the learning rate. Since this is a fixed learning rate within each block, we can apply the logarithmic-time EXP3 of Section~\ref{sec: log_time_exp3} or the constant-time EXP3 of Sections~\ref{sec: constant_time_exp3} and~\ref{sec: simpler_constant_time_exp3}. Although the data structure needs to be reconstructed with a new learning rate every $K$ rounds, the per-round computational complexity remains $O(1)$. In particular, Section~\ref{sec: simpler_constant_time_exp3} already reconstructs the data structure every $K$ rounds anyway, so no additional computational cost is incurred.

This approach parallels the doubling trick by employing a constant learning rate within each segment. However, whereas the doubling trick suffers from a delay on the order of the horizon, our method reduces this to only $K$ rounds---a much finer granularity that keeps the regret constant smaller. Furthermore, unlike the doubling trick which effectively resets and loses information at segment boundaries, this approach retains the learning history. This continuity leads to greater stability and is expected to yield superior practical performance.

\begin{proposition} \label{prop: anytime-slow-update-regret}
    The regret upper bound of constant-time EXP3 with delayed parameter updates is \begin{equation}
        \bar{R}_T \le 2 \sqrt{K T \ln K} + K \sqrt{\ln K} \label{eq: anytime-slow-update-regret}
    \end{equation} That is, the leading term depending on $T$ is the same as that of exact anytime EXP3.
\end{proposition}

\begin{tcolorbox}[colframe=gray!10,colback=gray!10,sharp corners,breakable]
\begin{proof}
    From the standard analysis of EXP3~\cite[Theorem 3.1]{bubeck2012regret}, the expected pseudo-regret of EXP3 with non-increasing learning rate $\eta_t$ is bounded by \begin{equation}
        \bar{R}_T \le \frac{\ln K}{\eta_T} + \frac{K}{2} \sum_{t=1}^T \eta_t
    \end{equation} Therefore, the expected pseudo-regret of EXP3 with delayed parameter updates is bounded as follows.
\begin{align}
    \bar{R}_T &\le \frac{\ln K}{\eta_{\tau(T)}} + \frac{K}{2} \sum_{t=1}^T \eta_{\tau(t)} \\
    &\le \frac{\ln K}{\sqrt{\frac{\ln K}{K \tau(T)}}} + \frac{K}{2} \sum_{t=1}^T \sqrt{\frac{\ln K}{K \tau(t)}} \\
    &= \sqrt{K \tau(T) \ln K} + \frac{\sqrt{K \ln K}}{2} \sum_{t=1}^T \sqrt{\frac{1}{\tau(t)}} \\
    &\le \sqrt{K \tau(T) \ln K} + \frac{\sqrt{K \ln K}}{2} \sum_{t=1}^T \sqrt{\frac{1}{t}} \\
    &\le \sqrt{K \tau(T) \ln K} + \frac{\sqrt{K \ln K}}{2} (2 \sqrt{T}) \\
    &= \sqrt{K T \ln K} + \sqrt{K \tau(T) \ln K} \\
    &\le \sqrt{K T \ln K} + \sqrt{K (T + K) \ln K} \\
    &\le \sqrt{K T \ln K} + \sqrt{K T \ln K} + \sqrt{K^2 \ln K} \\
    &= 2 \sqrt{K T \ln K} + K \sqrt{\ln K}.
\end{align}
\end{proof}
\end{tcolorbox}

Using the logarithmic-time EXP3 of Section~\ref{sec: log_time_exp3}, this expected pseudo-regret can be achieved in $O(\log K)$ worst-case time per round. Using the constant-time EXP3 of Sections~\ref{sec: constant_time_exp3} and~\ref{sec: simpler_constant_time_exp3}, this expected pseudo-regret can be achieved in expected $O(1)$ time per round. These improve upon the method of \citet{chewi2022rejection}, which achieves expected pseudo-regret of $4 \sqrt{K T \ln K}$ in $O(\log^2 K)$ time, in both computational time and regret.

\section{Conclusion}

In this paper, we proposed computationally efficient methods for EXP3. We proposed logarithmic-time EXP3 and constant-time EXP3, and showed that both improve upon previous methods in terms of both computational time and regret. We also demonstrated that computational efficiency can be achieved while preserving the leading term of the regret even when the horizon is unknown.

\bibliography{main}
\bibliographystyle{abbrvnat}

\appendix

\end{document}

%% file: math_commands.tex
\usepackage{amsmath,amsfonts,bm}

\def\eqref#1{equation~\ref{#1}}
\def\1{\bm{1}}

\DeclareMathAlphabet{\mathsfit}{\encodingdefault}{\sfdefault}{m}{sl}
\SetMathAlphabet{\mathsfit}{bold}{\encodingdefault}{\sfdefault}{bx}{n}

%% file: main.bib
@inproceedings{hazan2011simple,
  author       = {Elad Hazan and
                  Satyen Kale},
  title        = {A simple multi-armed bandit algorithm with optimal variation-bounded
                  regret},
  booktitle    = {Proceedings of the 24th Annual Conference on Learning Theory, {COLT}},
  volume       = {19},
  pages        = {817--820},
  year         = {2011},
  url          = {http://proceedings.mlr.press/v19/hazan11b/hazan11b.pdf},
}

@misc{koolen2024mltlecture9,
  author       = {Wouter M. Koolen},
  title        = {Machine Learning Theory 2024, Lecture 9},
  year         = {2024},
  url          = {https://homepages.cwi.nl/~wmkoolen/MLT_2024/slides9.pdf},
  note         = {Accessed: 2025-12-02}
}

@article{auer2002nonstochastic,
  author       = {Peter Auer and
                  Nicol{\`{o}} Cesa{-}Bianchi and
                  Yoav Freund and
                  Robert E. Schapire},
  title        = {The Nonstochastic Multiarmed Bandit Problem},
  journal      = {{SIAM} Journal on Computing},
  volume       = {32},
  number       = {1},
  pages        = {48--77},
  year         = {2002},
}

@inproceedings{chewi2022rejection,
  author       = {Sinho Chewi and
                  Patrik R. Gerber and
                  Chen Lu and
                  Thibaut Le Gouic and
                  Philippe Rigollet},
  title        = {Rejection sampling from shape-constrained distributions in sublinear
                  time},
  booktitle    = {International Conference on Artificial Intelligence and Statistics,
                  {AISTATS}},
  volume       = {151},
  pages        = {2249--2265},
  year         = {2022},
}

@article{matias2003dynamic,
  author       = {Yossi Matias and
                  Jeffrey Scott Vitter and
                  Wen{-}Chun Ni},
  title        = {Dynamic Generation of Discrete Random Variates},
  journal      = {Theory of Computing Systems},
  volume       = {36},
  number       = {4},
  pages        = {329--358},
  year         = {2003},
}

@article{bubeck2012regret,
  title={Regret analysis of stochastic and nonstochastic multi-armed bandit problems},
  author={Bubeck, S{\'e}bastien and Cesa-Bianchi, Nicolo and others},
  journal={Foundations and Trends{\textregistered} in Machine Learning},
  volume={5},
  number={1},
  pages={1--122},
  year={2012},
  publisher={Now Publishers, Inc.}
}

@article{walker1977efficient,
  author       = {Alastair J. Walker},
  title        = {An Efficient Method for Generating Discrete Random Variables with
                  General Distributions},
  journal      = {{ACM} Trans. Math. Softw.},
  volume       = {3},
  number       = {3},
  pages        = {253--256},
  year         = {1977},
}

@book{cesa2006prediction,
  title={Prediction, learning, and games},
  author={Cesa-Bianchi, Nicolo and Lugosi, G{\'a}bor},
  year={2006},
  publisher={Cambridge university press}
}
